\documentclass[letterpaper, 10 pt, conference]{ieeeconf}  % Comment this line out if you need a4paper

\IEEEoverridecommandlockouts                              % This command is only needed if 
\usepackage{xcolor}
\usepackage{subfig}
\usepackage{graphicx}
\usepackage{lipsum}
\usepackage{listings}% use for inserting a snipet of code in the manuscript
\usepackage{color}
\usepackage{mathtools}

\usepackage{graphicx}% http://ctan.org/pkg/graphicx
\usepackage{yhmath}% http://ctan.org/pkg/yhmath
\usepackage{mathdots}% http://ctan.org/pkg/mathdots
\usepackage{MnSymbol}% http://ctan.org/pkg/mnsymbol

\newcommand{\RizE}[1]{\textcolor{black}{#1}} % Riza' edits: this can be used to add your text to the manuscript.
\title{\LARGE \bf
Natural Language Robot Programming: \\NLP integrated with autonomous robotic grasping 
}

\author{Muhammad Arshad Khan$^{1\dagger}$, Max Kenney $^{1\dagger}$, Jack Painter $^{1\dagger}$, Disha Kamale$^3\dagger$, \\ Riza Batista-Navarro$^2$, Amir Ghalamzan E.$^{1\dagger}$% <-this % stops a space
\thanks{$^\dagger$ These authors have equivalent contribution.}%
\thanks{$^1$University of Lincoln, UK}
\thanks{{\tt\small aghalamzanesfahani@lincoln.ac.uk}}%
\thanks{$^2$ University of Manchester, UK}
\thanks{$^3$ Lehigh University, USA}
}

\begin{document}

\maketitle
\thispagestyle{empty}
\pagestyle{empty}

%%%%%%%%%%%%%%%%%%%%%%%%%%%%%%%%%%%%%%%%%%%%%%%%%%%%%%%%%%%%%%%%%%%%%%%%%%%%%%%%
\begin{abstract}
In this paper, we present a grammar-based natural language framework for robot programming, specifically for pick-and-place tasks. Our approach uses a custom dictionary of action words, designed to store together words that share meaning, allowing for easy expansion of the vocabulary by adding more action words from a lexical database. We validate our Natural Language Robot Programming (NLRP) framework through simulation and real-world experimentation, using a Franka Panda robotic arm equipped with a calibrated camera-in-hand and a microphone. Participants were asked to complete a pick-and-place task using verbal commands, which were converted into text using Google's Speech-to-Text API and processed through the NLRP framework to obtain joint space trajectories for the robot. Our results indicate that our approach has a high system usability score. The framework's dictionary can be easily extended without relying on transfer learning or large data sets. In the future, we plan to compare the presented framework with different approaches of human-assisted pick-and-place tasks via a comprehensive user study.

\end{abstract}

%%%%%%%%%%%%%%%%%%%%%%%%%%%%%%%%%%%%%%%%%%%%%%%%%%%%%%%%%%%%%%%%%%%%%%%%%%%%%%%%
\section{INTRODUCTION}

% STORY: various ways explored to assist humans in everyday and labour demanding tasks - some of which include haptic feedback, etc (give more ex) - but there is a learning curve to it- may not be very intuitive for layman - In this paper, we try to address a question as to how can these interactions be made more natural for humans without compromising on robot's safety and performance.}
% 1. Need for cobots - applications \\
% 2. safety and performance enhancing features \\
% 3. Our bridge - verbal communication natural\\
% 4. Contributions\\
The growing adoption of robotics automation in various industries has led to a need for humans and robots to share workspace, interact, and collaborate. However, commanding a robot to perform a task in such workspaces can be challenging. Although learning from demonstration, traditional teleoperated systems, and more recent collaborative robots provide useful tools for human-robot interaction, they still fall short of human-human interaction~\cite{colgate1996cobots, ajoudani2018progress}. Human workers use natural language to ask others to conduct a task for them while using teach pendants and joysticks can be difficult for untrained workers to communicate with the robot. Therefore, using natural language for intuitive collaboration between human and robot workers is crucial to unlocking the potential of bringing robots to human workspaces.

% With the rapid adoption of robotics automation in various industries, there is a growing need for humans and robots to share workspace, interact, and collaborate. However, commanding a robot to perform a task in such workspaces can be challenging. Learning from demonstration, traditional teleoperated systems, and more recent collaborative robots provide useful tools for human-robot interaction, but they are still far from human-human interaction. Human workers use natural language to ask others to conduct a task for them. Using teach pendants and joysticks can be difficult for untrained workers to tell the robot what to do. Therefore, using natural language for intuitive collaboration between human and robot workers is one of the key elements to unlocking the opportunity of bringing robots to human workspaces.

\begin{figure}[tb!]
\centering
\includegraphics[width=.8\linewidth]{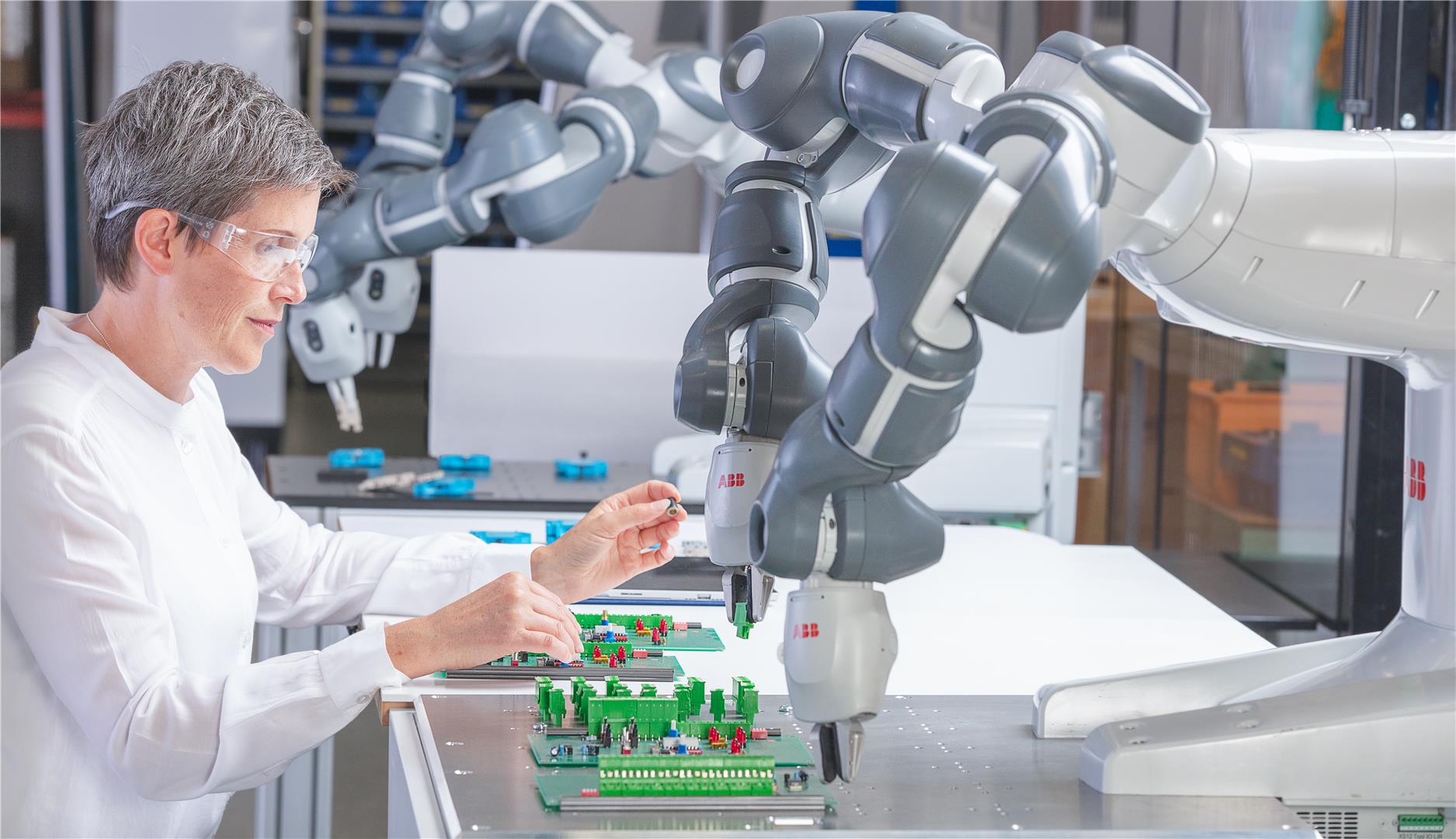}
\caption{YuMi® - IRB 14000 | Collaborative Robot~\cite{yumi} co-working with a human while the tasks are shared between the cobot and human. The safety features (which may include robot compliance, intention recognition, emergency bottoms, continuous monitoring of the robot and its workspace state via force/torque sensors, several visual sensing devices, proximity, and laser scanner sensors) allow secure task completions.}
\label{fig::yumi}
\vspace{-5pt}
\end{figure}

Fig.\ref{fig::yumi} illustrates a collaborative scenario involving a Yumi cobot \cite{yumi} from ABB and a human performing complementary tasks that may involve cognitively demanding tasks by the human and the precision ability of the robot worker. In order to ensure safe and efficient interactions, cobots are equipped with various features, including compliance, power, and force limiting, and haptic feedback. These features enable cobots to operate closely with humans in applications such as care robots\cite{van2013designing}, assistive robots~\cite{bemelmans2012socially}, and museum guide robots~\cite{del2019lindsey}. Additional features, such as continuous monitoring of the robot and its environment state through force, torque, visual, proximity, and laser scanner sensors, further improve the cobot's efficacy. However, these approaches are still not as effective as human-human co-working~\cite{demir2019industry}. Although beneficial to human operators, these approaches are associated with a limited capacity of human-robot and have a learning curve that may hinder their immediate usability. This paper aims to explore how natural language can be used to facilitate human-robot interactions in a more intuitive manner without compromising robot safety and performance.

% Fig.~\ref{fig::yumi} shows a sample collaborative scenario involving a Yumi cobot \cite{yumi} (manufactured by ABB) and a human where the robot can perform a set of precision operations while the human performs some less precise but cognitively more challenging tasks which showcase the complementary competence of a human and a robot. To ensure safe, secure and efficient interactions, these robots are equipped with many features such as compliance, power and force limiting, haptic feedback, etc. These features enable the cobots to work at close quarters with humans, e.g., care robots~\cite{van2013designing}, and assistive robots~\cite{bemelmans2012socially}, museum guide robots~\cite{del2019lindsey}, etc. Moreover, features like continuous monitoring of the robot as well as its environment state via force/torque/visual/proximity/laser scanner sensors have been implemented to further enhance efficacy. However, these approaches are still not as efficient as human-human co-working~\cite{demir2019industry}. Although immensely helpful to human operators, they usually have a learning curve associated with them and thus, may not be immediately usable to the end-users. In this paper, we aim to address the question of how human-robot interactions can be made more natural for humans without compromising on robot safety and performance. 
% <<<<<<<<<< >>>>>>>>>>>
% <<<<<<<<<< >>>>>>>>>>>

Verbal communication plays a crucial role in facilitating natural interactions among human workers and coordinating task executions within a team. This aspect serves as a key differentiator between human-human and human-robot workers. In recent years, researchers have explored the use of verbal communication for coordination between mobile robots and humans, as evidenced by studies such as~\cite{hameed2016using}. One such application is in tour guiding, where verbal communication is primarily used to provide the necessary information to the human. Additionally, Zhou et al. (2021)\cite{zhou2021inverse} proposed a novel approach for training a reinforcement learning agent using natural language goal-conditioned adversarial learning algorithm and a variational goal generator to sample natural language goals from trajectories of given demonstrations.

% Verbal communication would facilitate such natural interactions and is extensively used to coordinate task executions among members of a team of human workers. 
%
% This is a major differentiating factor between human-human and human-robot workers.
%
% A line of research studies the use of verbal communication for coordination between mobile robots and humans~\cite{hameed2016using}, e.g., in tour guiding where the verbal communication is mostly used to help the human with necessary information. Similarly \cite{zhou2021inverse} proposed to train a reinforcement learning agent with natural language goal-conditioned adversial learning algorithm and a variational goal generator to sample the NL goals from trajectories of given demonstrations. 

%
%While a very limited body of research has been devoted to the use of verbal commands for mobile/social robots~\cite{hameed2016using}, \RizE{natural language processing} (NLP) has been unexplored for coordinating task execution between a manipulating robot and a human. 
%

% Employing natural language processing (NLP) for coordinating task execution between a manipulating robot and a human, is more challenging than the use of NLP for mobile/service robots as in contrast to the latter, a manipulator is expected to physically interact with its environment with fine-grained actions, e.g. to reach an object and form stable grasp/contacts with it and deliver it at the target location~\cite{mavrakis2017safe}.\\  
%

Coordinating task execution between a manipulating robot and a human using natural language processing (NLP) is a more challenging task than using NLP for mobile/service robots. This is because, unlike mobile/service robots, manipulator robots are expected to physically interact with their environment through precise and fine-grained actions, such as reaching for an object, forming stable grasp or contacts with it, and delivering it to the target location~\cite{mavrakis2017safe}.

This paper introduces one of the first frameworks for verbally commanding a cobot to perform pick-and-place tasks. Pick-and-place is a common yet straightforward task to automate. The proposed framework comprises three key components: natural language processing, semantic segmentation, and autonomous grasping. The autonomous grasping module generates grasping poses for a given scene, allowing a human operator to command the robot to execute a desired grasp. The contribution of this paper is manifold: (i) We created a dictionary of the common tasks involved in robotic manipulation in our NLP component. (ii) Our framework includes a translator-middleware capable of translating natural language instructions into low-level joint space or operational space robot commands. (iii) We provide two-level verbal commands in our framework, i.e., high (task) level and low (trajectory) level (either joint or operational space trajectory). (iv) We present a unified framework for verbally commanding a manipulator, integrating NLP with autonomous grasping.
Finally, we demonstrate the effectiveness of our proposed framework in a series of experiments.
\section{Related Work}\label{related_works}
% \paragraph{Robotic Grasping}
Synthesising grasp poses from single or multiple point clouds is a topic of significant research interest. Several approaches have been proposed, including sim-to-real learning~\cite{james2019sim}, deep learning via domain randomisation~\cite{tobin2018domain}, learning hand-eye coordination for grasping unknown objects~\cite{levine2018learning}, and probabilistic generative models of grasping configurations~\cite{kopicki2019learning}. 
%Researchers have found that shape completion can improve the performance of synthesizing grasping configurations for unknown objects, such as through primitive shape fitting to point cloud~\cite{baronti2019primitive} as reported by~\cite{varley2017shape}. However, autonomous grasping approaches have lacked reliable robustness for conservative industries such as nuclear waste decommissioning. For example, Kopicki et al. \cite{kopicki2019learning} proposed an approach that can be partially generalized across different working conditions, unknown objects, and unknown environments. Although their approach improved performance reported in \cite{kopicki2019learning}, it still does not meet the success rate demanded by conservative industries~\cite{talha2016towards}. 
%Recent advances in deep neural networks allowed for further improvement of grasping algorithms. %For instance, Mousavian et al.~\cite{mousavian20196} proposed a deep neural network that generates grasping configurations for a parallel jaw gripper given a point cloud scene and evaluates the quality of each grasp to determine the optimal one. 
A comprehensive review of available deep learning methods for grasp synthesis is presented in~\cite{newbury2022deep}.

% \pragraph{{Language Grounding}}
The process of correlating entities referred to in a natural-language expression with an agent's physical environment is known as language grounding~\cite{misra2016tell}. Referring expression comprehension is a type of language grounding that focuses on recognising the object in an image referred to in a natural-language expression \cite{qiao2021}. This problem has been addressed using deep learning-based approaches. For example, Mao et al.~\cite{mao2016generation} used a CNN-LSTM network to identify the image region with the highest probability of being the target object. Lu et al.~\cite{lu2019vilbert} used a transformer-based ViLBERT model to jointly learn language and image representations and select the most likely target object.

Shridhar et al.~\cite{shridhar2020alfred} present an approach of learning long-horizon tasks, which consist of action sequences such as washing a coffee mug, represents a much more advanced form of language grounding. State-of-the-art language models have been employed to predict the granular actions an agent needs to take for learning such tasks \cite{suglia2021embodied,ahn2022can}. In our work, we focus on executing short-horizon tasks that require a robot to execute only one action at a time. Nevertheless, our work goes beyond the recognition of referring expressions (which is only one component of our work) since our agent executes a pick-and-place task within its physical environment through robotic manipulation.

Various methods have been proposed for enabling robots to understand natural language commands, which can be broadly categorised into two types: (1) supervised and (2) grammar-based approaches. Kollar et al.\cite{kollar2010toward} proposed the use of spatial description clauses (SDCs) to transform natural language commands into semantic structures, with each SDC comprising four components: an event, an object, a place, and a path. A conditional random fields (CRF) model was trained to recognise these four components in a given natural language command. However, the main drawback of supervised approaches, such as SDC transformation, is their reliance on labelled data. For example, Tellex et al.\cite{tellex2011understanding} used SDC transformation for robotic forklift manipulation but required a large manually labelled corpus of commands constructed through crowdsourcing.

On the other hand, grammar-based approaches do not require labelled training data, although they need constituency parsers to convert natural language instructions to a formalism that robots can understand, such as Robot Control Language (RCL). For instance, Matuszek et al.~\cite{matuszek2013learning} employed a combinatory categorical grammar (CCG) parser, while Packard 2014 used a probabilistic context-free grammar (PCFG) parser to generate RCL instructions. These grammar-based methods, however, may still have limitations in their ability to understand and process complex natural language commands.

% Methods which have been proposed for understanding natural-language commands or instructions for robots, can be broadly categorised into: (1) supervised and (2) grammar-based approaches. Kollar et al.~\cite{kollar2010toward} proposed the transformation of natural language commands into semantic structures called spatial description clauses (SDCs). Each SDC consists of four components: (i) an event, (ii) an object, (iii) a place and a (iv) path. A conditional random fields (CRF) model was trained to recognise these four components in a given natural language command. The drawback of such an approach, however, as with many supervised methods, is the reliance on labelled data. The approach of Tellex et al.~\cite{tellex2011understanding} for example, which made use of SDC transformation for robotic forklift manipulation, required a large manually labelled corpus of commands that was constructed using crowdsourcing. Meanwhile, grammar-based approaches do not rely on labelled training data although they require constituency parsers for converting natural language instructions to a formalism that can be understood by robots, e.g., Robot Control Language (RCL). Matuszek et al.~\cite{matuszek2013learning} for instance, employed a combinatory categorical grammar (CCG) parser, while Packard 2014 used a probabilistic context-free grammar (PCFG) parser in order to generate RCL instructions. 

% \paragraph{Natural language for robot planning} 
Natural language is proposed in a framework for correcting robot planning~\cite{sharma2022correcting}. Suglia et al.\cite{suglia2021embodied} presented a language-guided visual task completion method for interactive navigation tasks. Ahn et al.~\cite{ahn2022can} used pre-trained skills to provide real-world grounding and constrain the model to propose natural language actions that are both feasible and contextually appropriate.

%In this study, we also transform natural-language instructions into a formalism, namely, SDCs, but with a simple approach. Unlike Tellex et al.\cite{tellex2011understanding}, our NLP approach (described in the next section) requires little labelled training data. Moreover, we opted to use dependency parsers instead of constituency parsers, as used by Matuszek et al.~\cite{matuszek2013learning}. Dependency parsers enable us to extract semantic relations between words based on the output dependency tree directly, while constituency parsers require additional post-processing to extract semantic relations. Our approach only relies on an off-the-shelf dependency parser to transform natural language instructions into SDC structures.

%Sharma et al.~\cite{sharma2022correcting} presented a framework in which natural language is utilised to correct the robot planning. \cite{suglia2021embodied} presents a language-guided visual task completion for interactive navigation tasks. Ahn et al.~\cite{ahn2022can} proposed to provide real-world grounding by means of pretrained skills, which are used to constrain the model to propose natural language actions that are both feasible and contextually appropriate. 
In this work, we also convert natural-language instructions into a formalism, namely, SDCs, but using a simple approach. In contrast to the work by Tellex et al.~\cite{tellex2011understanding}, our NLP approach (described in the next section) does not require huge amounts of labelled training data. Furthermore, we chose to employ dependency parsers  rather than constituency parsers which were used by Matuszek et al. \cite{matuszek2013learning}. Dependency parsers allow us to directly extract semantic relations between words based on the output dependency tree, whilst constituency parsers produce output that still require post-processing if we wish to extract semantic relations. Our approach relies only on an off-the-shelf dependency parser for transforming natural language instructions into SDC structures.

%\textcolor{red}{: could you please differentiate our work wrt~\cite{ralph2008toward,rao2018learning, hatori2018interactively, tellex2011understanding, matuszek2013learning} and other relevant ones? You may briefly describe the NLP side of the framework using Fig.~\ref{fig:deptree}}.

% <<<<<<<<<<<Fig>>>>>>>>>>>
% \begin{figure}[tb!]
% \centering
% \includegraphics[width=.99\columnwidth]{images/new_images/VSC_Framework_Architecture.png}
% \caption{{\small An overview of {\disha{VSC for}} robotic manipulation tasks.}}
% \label{fig:vsc}
% \vspace{-0.0cm}
% \end{figure}
% <<<<<<<<<<<Fig>>>>>>>>>>>

\section{Natural Language Robot Programming}
%\subsection*{Autonomous grasping}
% \input{parts/grasp}

Simple pick-and-place is one of the major tasks for robotic manipulators yet not fully solved. 
Technology of autonomous grasping for simple pick-and-place task is maturing~\cite{newbury2022deep}. 
Autonomous grasping refers to an algorithms compute one or several grasping configuration for a manipulator given the visual sensory data, e.g. RGB-D data. 
In contrast to the classical analytical grasping approach (which computes necessary contacts/forces between object and robot's fingers to make an equilibrium which is stable to some extends against some external forces~\cite{bicchi2000robotic}).

\begin{figure}[tb]
\centering
\includegraphics[width=.9\columnwidth, trim= 0cm 0cm 0cm 0cm, clip]{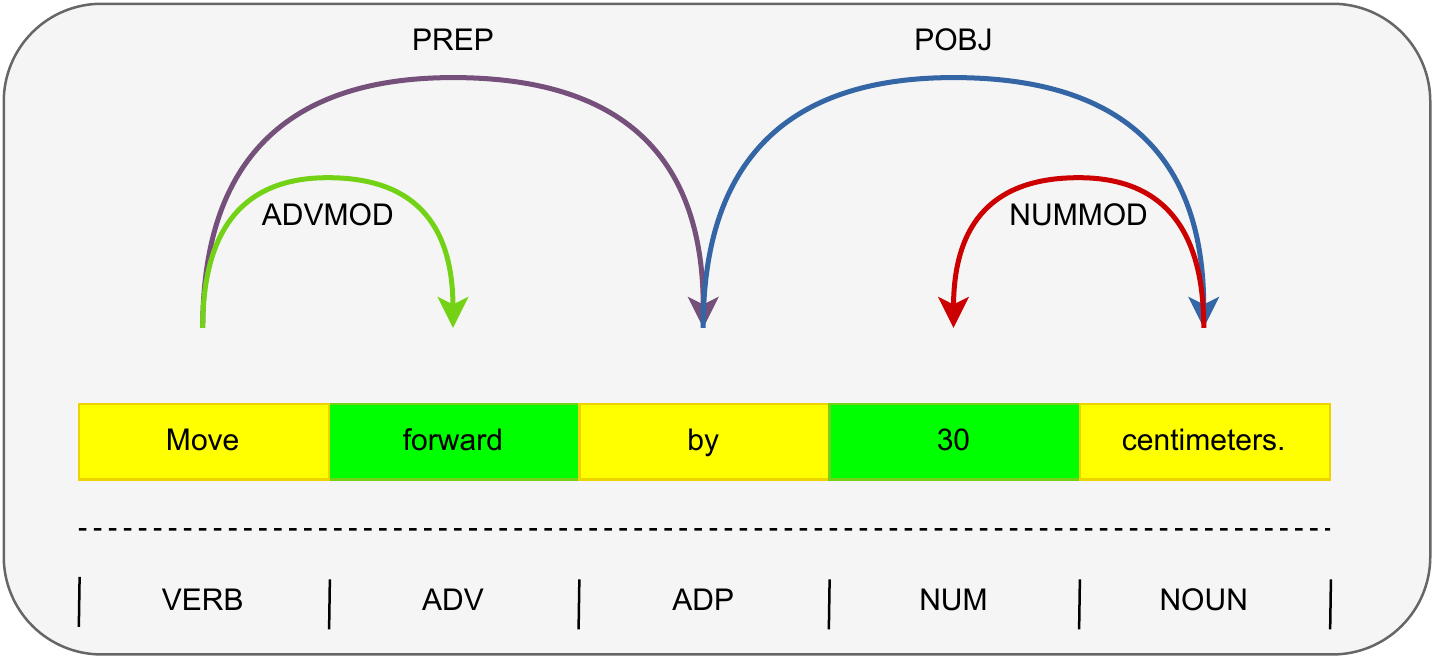}
\caption{ Dependency tree generated by \textit{spaCy} visualised. Each arrow shows a relationship between two words. Here it identified 'Move' as the main verb and that it has as its dependants  'forward' which is an adverb, and 'by' which is a preposition. By traversing through the dependencies one is also able to reach '30', a number since it is a dependent of 'centimetres', a noun. This system of hopping through dependencies and tracking part-of-speech tags allows us to create rules which generate the final action command. } 
\label{fig:deptree}
\vspace{-.5cm}
\end{figure}
% <<<<<<<<<<<Fig>>>>>>>>>>>

% \subsection*{Natural Language Processing}
% \input{parts/nlp}
% \subsubsection{High (task) level SDC: }
% \textbf{Speech module of VSC:} 

\subsection{Speech module of NLRP}
While we integrate a simple grasping in our Natural Language Robot Programming (\textbf{NLRP}) module (discussed in the next section), integration of available state-of-the-art grasping methods in our NLRP is straightforward. Our approach to understanding natural language commands is underpinned by a dependency parser, specifically the implementation made available by \textit{spaCy} \cite{honnibal-johnson:2015:EMNLP}. Dependency parsing automatically extracts the syntactic structure of a sentence as a tree, by identifying typed binary relationships between its words \cite{jurafsky2000speech}. 

\begin{figure}[tb!]
	\centering		
	\subfloat[][High level SDC]{\includegraphics[width=0.9\columnwidth, trim=.01cm 0cm .01cm 0cm ,clip]{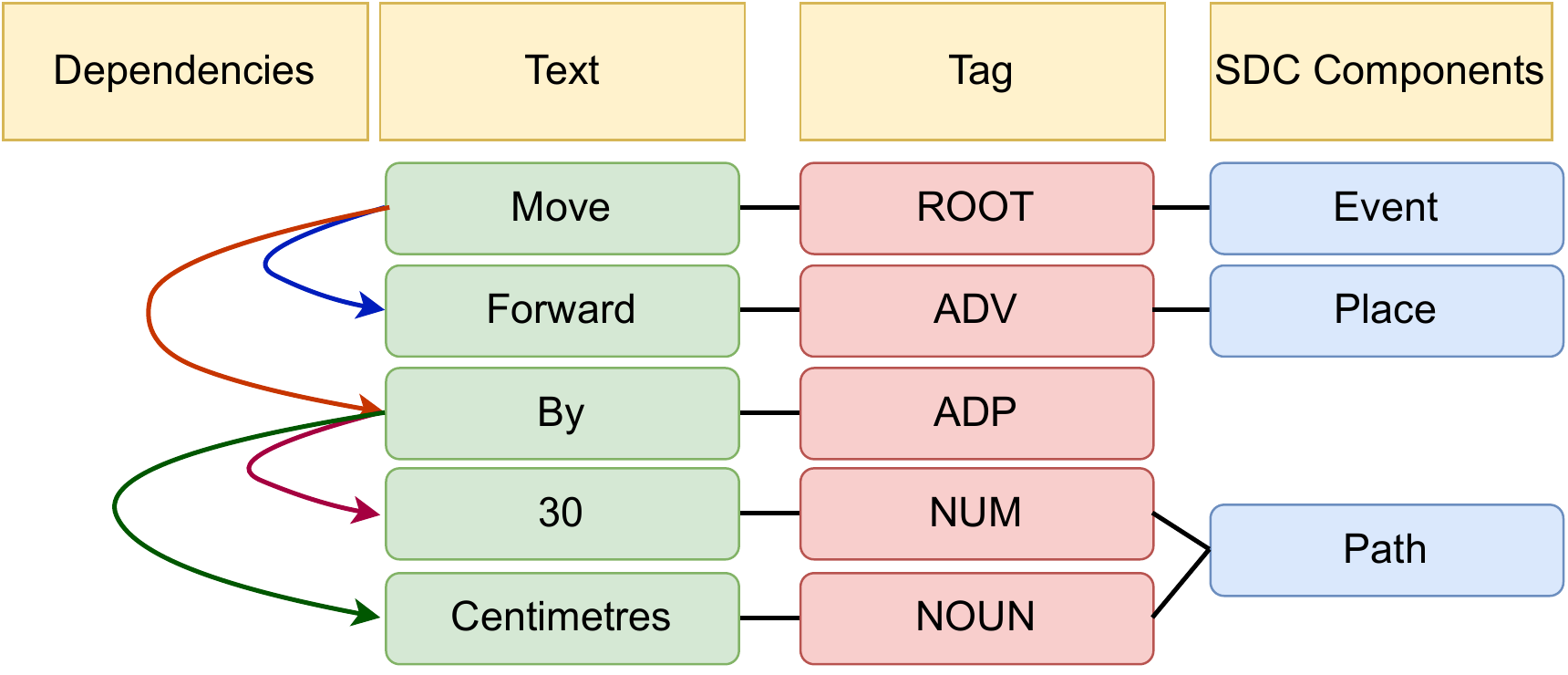}}
 	\vspace{-.0cm}\\
	% \subfloat[][VSC Network Schematic]{\includegraphics[width=0.9\columnwidth, trim=2cm 8cm 2cm 8cm ,clip]{images/High_Level_System_Diagram.pdf}}
 	% \vspace{-.0cm}
\caption{ Diagram visualising how the input text corpus is transformed into SDC format to output the final robot action using each words tag and dependencies sourced from \textit{spaCy} -- see Table~\ref{tab:vsc_dictionary}; }
\label{fig:nlp_sdc}
\vspace{-.0cm}
\end{figure}

Each of these binary relationships encodes a dependency between two words: a \emph{head} and its \emph{dependent} (a word that modifies or specifies the head). The head of the entire sentence - typically the main verb - becomes the root of the dependency tree. Each \emph{head-dependent} relationship is assigned a type, which is any of the known English grammatical functions (e.g., subject, direct object, indirect object, etc.). The \emph{head-dependent} relations are thus able to capture semantic relationships (e.g., between an action and a receiver of that action). In Fig.~\ref{fig:deptree}, we show a sample sentence together with its dependency tree, as extracted by spaCy. Each of the words in the sentence appears as a node with the part-of-speech (POS) tag indicated, specifying if the word is a noun, verb, or preposition (~Fig.~\ref{fig:nlp_sdc}). 

% [{\disha{TODO: description of fig 4}}]. 

% <<<<<<<<<<<Fig>>>>>>>>>>>

% <<<<<<<<<<<Fig>>>>>>>>>>>

\iffalse

% <<<<<<<<<<<Fig>>>>>>>>>>>
\begin{figure}[tb!]
\centering
\includegraphics[width=0.9\columnwidth, trim=2cm 8cm 2cm 8cm ,clip]{images/High_Level_System_Diagram.pdf}
\caption{{\small High level diagram showing the flow of data from the client side application to the robot. }}
\label{fig:dataflow}
\vspace{-.8cm}
\end{figure}
% <<<<<<<<<<<Fig>>>>>>>>>>>

% <<<<<<<<<<<Fig>>>>>>>>>>>
\begin{figure}[tb!]
\centering
\includegraphics[width=0.9\columnwidth, trim=2cm 4cm 2cm 3cm ,clip]{images/SDC_diagram.pdf}
\caption{{\small Diagram visualising the how the Natural Language Processing creates the format (SDC) to output the final robot action using each words tag and dependencies sourced from \textit{spaCy}.}}
\label{fig:sd}
\vspace{-.8cm}
\end{figure}
% <<<<<<<<<<<Fig>>>>>>>>>>>
\fi

%The information in the dependency tree obtained by \textit{spaCy} is then expressed in the form of a spatial description clause (SDC). As mentioned in the previous section, an SDC consists of four components, namely:
In our approach, we leverage the information encoded in the dependency tree generated by the \textit{spaCy} parser to construct a spatial description clause (SDC). As described in the previous section, an SDC comprises of four key components, namely:
\begin{itemize}
\item an \emph{event} (\emph{E}): the action that the robot is required to execute
\item an \emph{object} (\emph{O}): the item that a command is directed at (can be any other object in the world or even a part of the robot itself)
\item a \emph{place} (\emph{PL}): a place in the world such as a direction or the axes to move along
\item a \emph{path} (\emph{PA}): the extent (in terms of a specific amount) to which the action should be taken 
\end{itemize}
It is important to note that not all four components are always present in a natural language command. While the event component (\textit{E}) is mandatory, any of the remaining components can be unspecified depending on the command.
%We note that not all four above components are expressed in every command, except for event \emph{E}, any of them might be unspecified. 

As an initial step towards generating SDC structures from natural language instructions, we created a custom dictionary of action words specifically tailored for pick-and-place tasks. The dictionary was constructed in accordance with the grammar proposed by Song \cite{song2004natural}, which links high-level action words to other words that indicate similar actions. This enabled us to easily incorporate additional action words from the WordNet \cite{miller90wordnet} lexical database, and expand the vocabulary as needed. Our dictionary comprises seven action categories, each containing up to 17 high-level words that can be further expanded to include additional synonyms.

%\RizE{ As a first step towards SDC generation, we firstly compiled a custom dictionary of action words used in pick-and-place tasks. This dictionary was designed to accommodate storing together words that share meaning, according to the grammar proposed by Song \cite{song2004natural} in which high-level action words are linked to other words that signify similar actions.This allowed us to easily expand the vocabulary by adding more action words from the WordNet \cite{miller90wordnet} lexical database. In our final dictionary,  there are seven action categories, each with up to 17 high-level words that can further be expanded to store additional synonyms if needed. }

% <<<<<<<<<<<Fig>>>>>>>>>>>
% \begin{figure}[tb!]
% \centering
% \includegraphics[width= 1\columnwidth, trim= 1.5cm 3cm 1cm 3cm, clip]{images/Low_Level_System_Diagram_.pdf}
% \caption{{\small Low level diagram showing the responsibilities of each component as the data goes from client to robot.}}
% \label{fig:client}
% \vspace{-.0cm}
% \end{figure}
% <<<<<<<<<<<Fig>>>>>>>>>>>

\begin{table}[t!]
\vspace{1mm}
\hrule width \hsize \kern 1mm \hrule width \hsize height 2pt
\vspace{1mm}
\caption{SDC component shown in Fig.~\ref{fig:nlp_sdc}. The baseline of the primitive action library of VSC. This library allows a human operator to amend/improve it by adding new words to the library. }
\vspace{-0mm}\hrule
\begin{center}
\begin{tabular}{c}
{\small
\begin{lstlisting}[mathescape]
//dictionary.js (*\bfseries {SDC} Component*)
class Dictionary {
 constructor(){
 // Synonyms of the verbs
 this.Verbs = {
  "Move": ["move", "go", "travel" $\displaystyle \cdots$],
  "Grab": ["grab", "grasp", "catch"  $\displaystyle \cdots$],
  "Rotate": ["rotate", "revolve" $\displaystyle \cdots$],
  $\displaystyle \vdots$
  };
  $\displaystyle \vdots$
  
 this.Objects = {
  "Hand": ['hand', 'fingers', 'wrist' $\displaystyle \cdots$],
  "TeddyBear": ['teddy bear', 'teddy' $\displaystyle \cdots$],
  "Bottle": ['water bottle', 'bottle' $\displaystyle \cdots$],
  "Scissors" : ["scissors", "scissor" $\displaystyle \cdots$], 
  $\displaystyle \vdots$
  };
  
 //Used for finding the Path
 this.PlaceWords = {
  "Forward": ["forward", "forwards", $\displaystyle \cdots$],
  "Backward": ["backward", "backwards", $\displaystyle \cdots$],
  "Up": ["up", "upward", "upwards", $\displaystyle \cdots$],
  $\displaystyle \vdots$ 
 };

 $\displaystyle \vdots$
 this.triggerWords = {
  "Learn": ["means", "implies" $\displaystyle \cdots$],
  "Split": ["then", "before" $\displaystyle \cdots$],
  "Stop": ["stop", "halt", "quit" $\displaystyle \cdots$]
  };
 }
}
\end{lstlisting}}\\

\end{tabular}
\end{center}
\vspace{0mm}
\hrule width \hsize \kern 1mm \hrule width \hsize height 2pt
\label{tab:vsc_dictionary}
\vspace{-0mm}
\end{table}

\RizE{
In almost all of the cases, the word at the root of the dependency tree (i.e., the main verb) is contained in our custom dictionary and becomes the event \emph{E} of the SDC. However, if the identified root is not in the dictionary, we iterate through all of the words in the sentence to find a match with the dictionary, which then becomes the value for \emph{E}.
}
\RizE{
Once \emph{E} is identified, we iterate through each of its dependent, and based on the part-of-speech tag assigned to a word, we perform a different process in order to populate the SDC. 
For example, within the context of our problem, a word tagged as `NNS' (plural noun) corresponds to a unit of measurement (e.g., ``centimeters''). 
When this is encountered, its dependencies are used to search for another word with the dependency relation tag `NumMod'; this word will be a number. 
By joining these two words together we can obtain a number and a unit of measurement which will make up the path \emph{PA} such as ‘30\_Centimetres’. 
A similar process is performed for both the object \emph{O} and place \emph{PL} of the SDC. %where they are additionally compared to the dictionary to see if a word found is a synonym of one of our high level words. This high level word is then stored as the place or path.
}

\RizE{
The resulting SDC is sent to the robot controller where depending on the high-level word found, it makes use of the other components to instruct the robot to perform the command. In real-time, this whole process takes seconds and allows for an easy and reactive control experience.
}

% <<<<<<<<<<<Fig>>>>>>>>>>>
\begin{figure}[tb!]
\centering
\includegraphics[width= 1\columnwidth, trim= 0cm 0cm 0cm 0cm, clip]{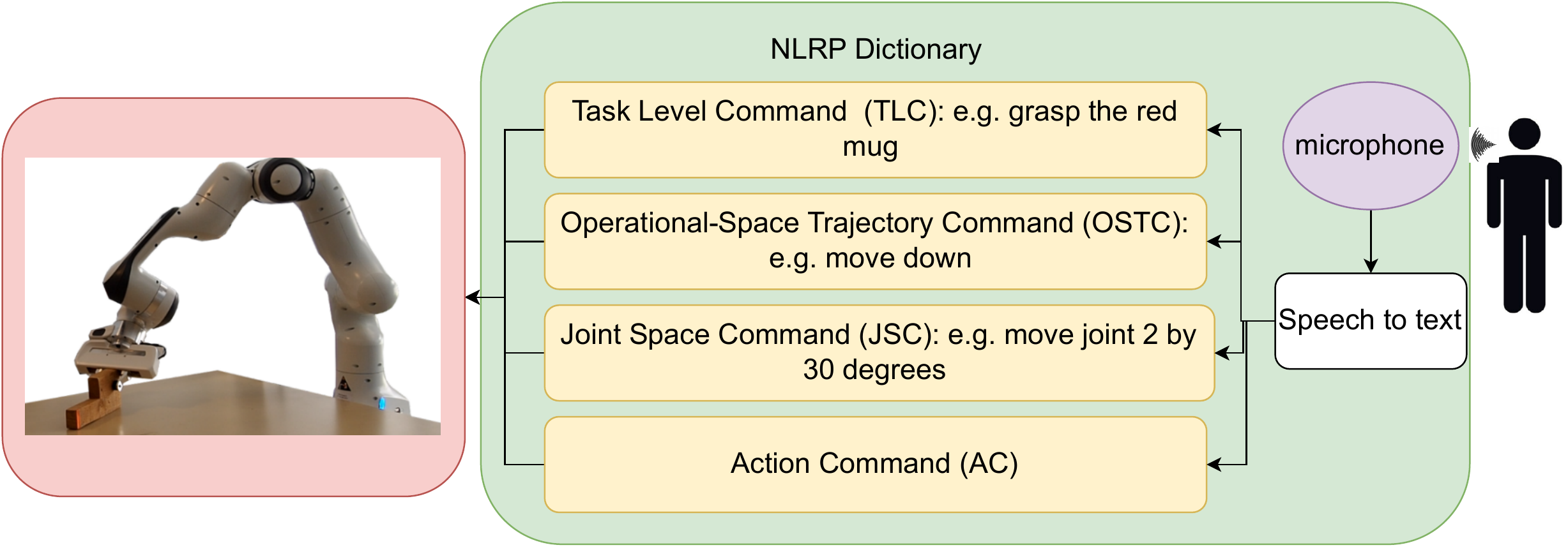}
\caption{Different levels of commanding the robotic manipulator in NLRP framework.}
\label{fig:low_lev_cont}
\vspace{-.4cm}
\end{figure}
% <<<<<<<<<<<Fig>>>>>>>>>>>

\subsection{Robot Controller Module of NLRP} 

Four different ways were designed to convert the speech commands in the form of a single SDC into a motion command for different manipulation tasks as shown in Fig.~\ref{fig:low_lev_cont}. These include:
% as per Fig.~\ref{fig:low_lev_cont}: 

\begin{enumerate}
    \item \textit{Task-level} - give the robot task-level commands such as \textit{grasp an object in the scene}.
    \item \textit{Operational-space Level} - command the manipulator to follow a specific path in operational-space such as \textit{move forward 30 cm}.
    \item \textit{Joint-space Level} - move a specific joint such as \textit{move joint 3 by +/- 15 degrees}.
    \item \textit{Action Level} - such as \textit{stop execution}, \textit{start execution} or \textit{recover from error}.
\end{enumerate}

Given an SDC $C$,  $\mathcal{B}$ takes in the current state of the robot and the given $C$ and outputs a joint trajectory as follows:

\begin{equation}
     \theta(t) = \mathcal{B} ( S_{i}, C_{i} )
     \label{eq:transform_eq}
\end{equation}

The mapping function from SDC $C$ to the joint space trajectory is conditional based on the \textit{event (E)} in the given SDC $C$. For example, if the given SDC $C$ specifies a Task-level or Task-space level command, then the robot Cartesian planner and the object 2D pose, in case of Task-level command are used to generate the joint trajectory $\theta(t)$. On the other hand, if the given SDC $C$ specifies a Joint-space level command, then the current robot joint-space states and the command joint-space instructions within the given SDC $C$ are used to generate a robot joint trajectory.

Given that it is possible that the \textit{path (PA)} may be missing from the given SDC, a set of default values for the Task-space and Joint-space commands are used. If the given SDC specifies a \textit{path (PA)} then the default values are overridden.

% \subsection{\disha{Autonomous Grasp Generation:
% {\emph{The DexNet or GraspNet module description}}}}

% \subsection{\disha{Task Execution: GUI
% {\emph{ description }}}}

\section{Experiments}

\begin{figure} 
\centering
\includegraphics[width=0.90\columnwidth]{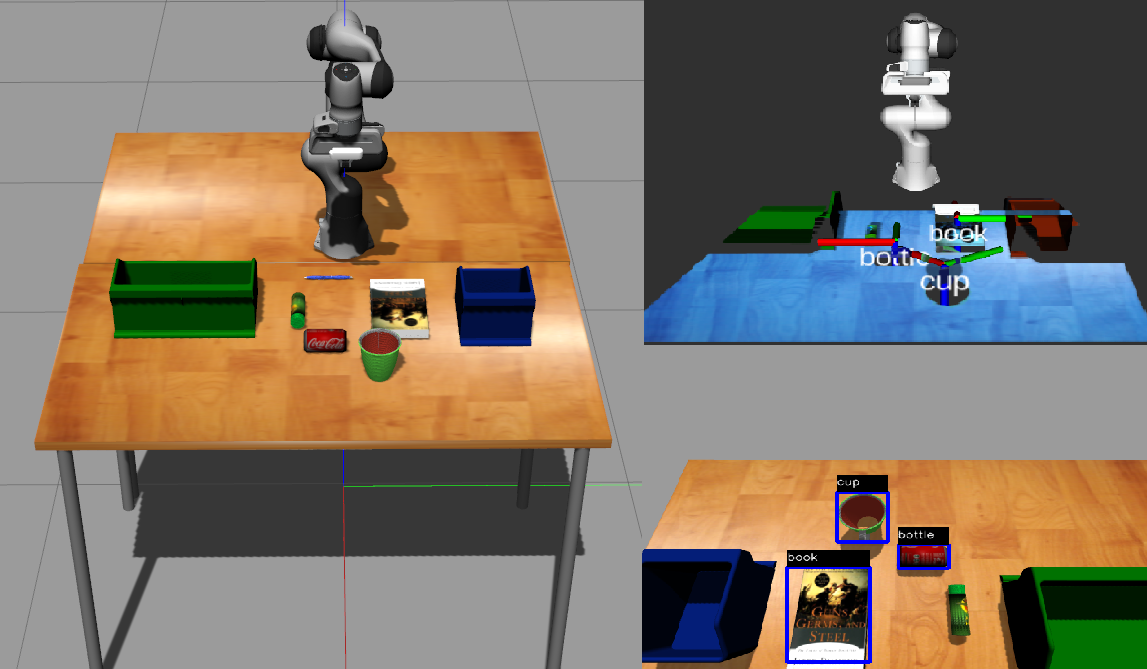}
     \caption{Experimental setup in simulation with a Franka arm on top of a table with multiple objects. 2D object grasp poses and semantic segmentation of each pick-able object are shown as well.}
	 \label{fig:simulated_ex_setup}
\end{figure}

\begin{figure}[tb!]
	\centering		
	\subfloat[][NLRP framework experimentation setup]{\includegraphics[width=0.9\columnwidth]{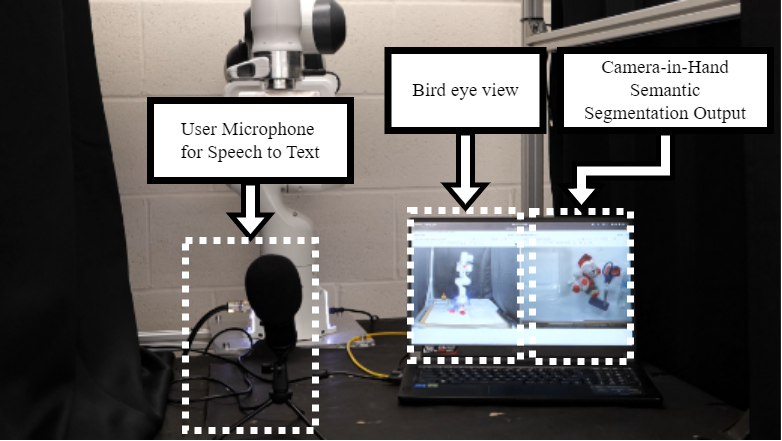}}
 	\vspace{-.0cm}\\
	\subfloat[][Semantic Segmentation Output]{\includegraphics[width=0.9\columnwidth]{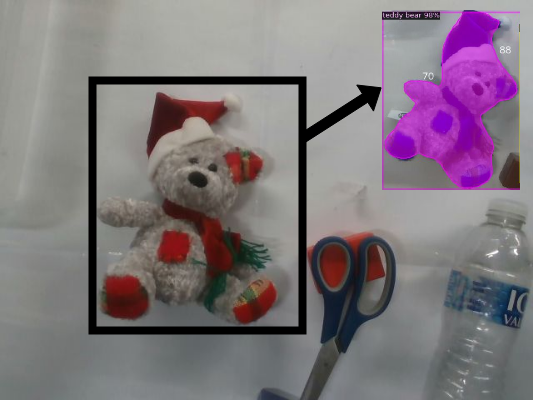}}
 	\vspace{-.0cm}\\

\caption{(a) Experimentation setup for the proposed NLRP framework with camera-in-hand and bird-eye view visual feedback to the user. (b) Mask-RCNN semantic segmentation output of the scene on the robot side.}

\label{fig:real_exp_setup}
\vspace{-.0cm}
\end{figure}

Our experimental setup for validating the proposed Natural Language Robot Programming (NLRP) framework consists of a 7-DoF Franka Panda robotic arm, manufactured by Franka Emika, equipped with a calibrated Intel Realsense D435i camera-in-hand. As illustrated in Fig.\ref{fig:simulated_ex_setup} and Fig.\ref{fig:real_exp_setup}, we conducted experiments in both the Gazebo simulation environment and the real world environment.

In our experiments, the robotic arm is placed on a table top in an environment that contains several common objects, such as a cup, book, bottle, and scissor, which are visible in the robot camera view. The camera-in-hand image-feed is processed by a Mask-RCNN model \cite{He_2017_ICCV}, which has been previously trained on the COCO dataset, to obtain semantic segmentation predictions, as shown in the bottom-right of Fig.\ref{fig:simulated_ex_setup} and Fig.\ref{fig:real_exp_setup} (b). The resulting semantic segmentation results are displayed on the user interface to inform users of the objects that are pick-able in the scene. Using the segmentation results and the calibrated depth image, we locate 2D grasp poses for each visible object, as demonstrated in the top-right of Fig.~\ref{fig:simulated_ex_setup}. To facilitate natural language interaction with the system, we compile a custom dictionary of objects, including their synonyms, and add them to the \textit{Objects}-category of our NLRP dictionary. For speech input, we use a Veetop USB Microphone to record the human operator's voice, which is then processed by the Google API.

% \begin{figure} 
% \centering
% \includegraphics[width=0.90\columnwidth]{images/new_images/Telepperation_Experimentation_Setup.png}
%      \caption{Teleoperation experimentation setup for the comparison user study with the same visual feedbacks as VSC framework experiments in Fig.~\ref{fig:real_exp_setup} (a). The leader arm is in controlled in gravity compensation control-mode with no haptic feedback-force control.}
% 	 \label{fig:teleop_ex_setup}
% \end{figure}

\section{Results \& Discussion} 

We evaluated the efficacy of our Natural Language Robot Programming (NLRP) framework in both task space and joint space through both simulation and real-world experimentation. To assess the effectiveness of NLRP, we conducted four different manipulation tasks involving the pick and place of two objects - a teddy bear and a water bottle - in any order, as shown in Fig.~\ref{fig:low_lev_cont}.

To further evaluate NLRP, we recruited nine participants who were provided with a brief description of the study and experiments. Participants utilized a microphone and two camera views, including a bird's eye view of the robot's workspace and a camera in hand. They were instructed to speak clearly and use any grammatically correct sentences they preferred to complete the task.

The participants' speech input from the microphone was converted to text corpus using Google's Speech to Text API. This corpus was then processed using our proposed NLRP-layered approach to obtain joint space trajectories for the different manipulation tasks. These trajectories were subsequently transmitted to the Franka Arm control through ROS controllers. After completing the pick and place task, each participant was requested to complete a questionnaire, which included evaluations of the System Usability Scale and NASA TLX~\cite{brooke1996sus}. This enabled us to assess the efficacy and usability of our proposed approach.

\begin{figure} 
\centering
\includegraphics[width=0.99\columnwidth]{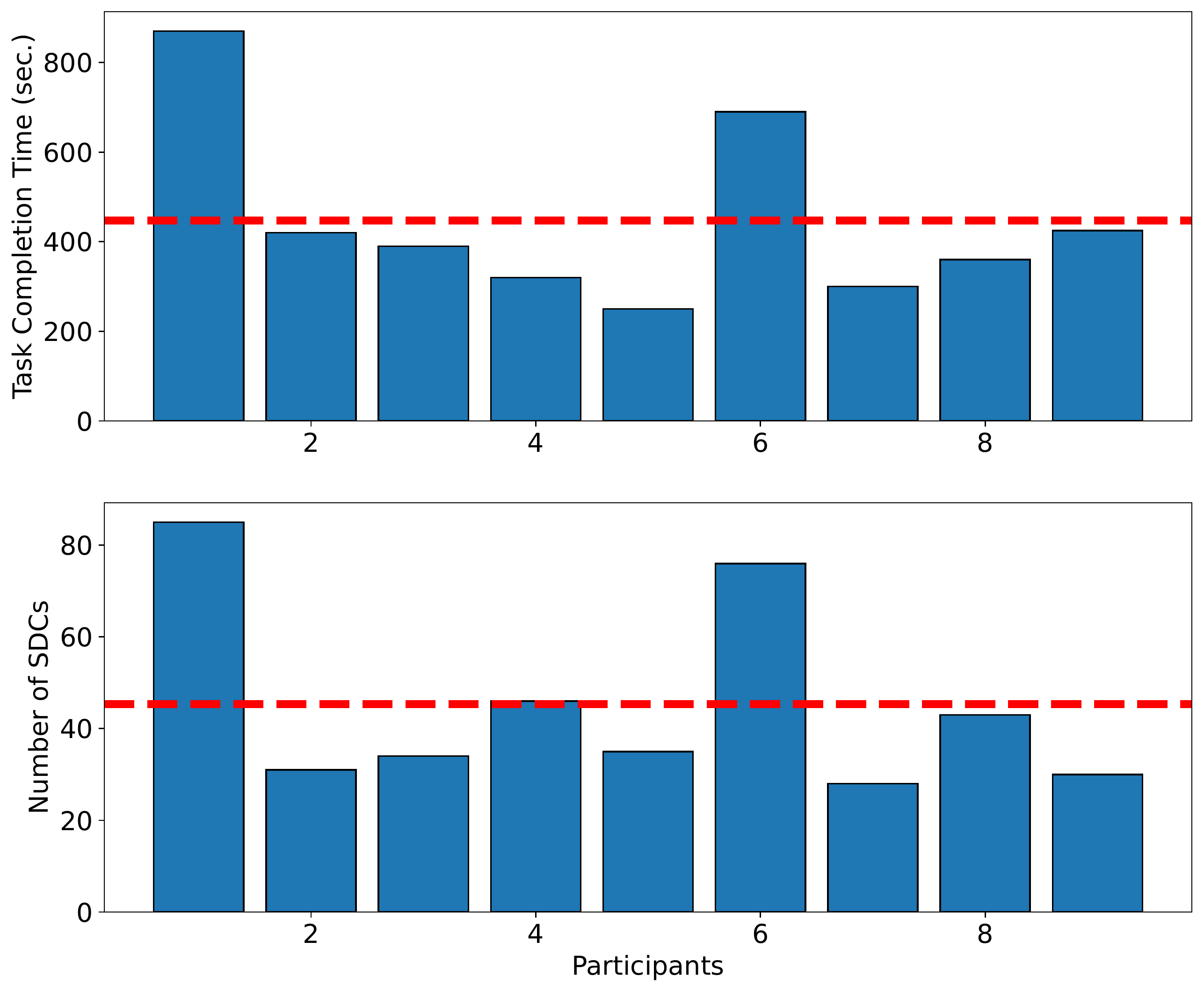}
     \caption{ Graphical representation of pick and place task completion time and number of SDCs used to complete the time. }
	      \label{result:plot_results}
\end{figure}

In our preliminary subject tests, 33\% of the participants were females and 67\% males. English is not the first language of participants. Fig.~\ref{result:plot_results} shows the number of SDC and time-to-complete the pick and place task. The average number of commands given by each participant was 45 and the average time in seconds taken to complete the pick and place task was 447.2 seconds as represented by the red lines on Fig.~\ref{result:plot_results}. The participant reported an average workload rating of 62 using NASA TLX cognitive workload assessment whereas an average of 58.7 system usability rating was reported using System Usability Scale.  

In future work, we plan to conduct comprehensive human-subject tests comparing the effectiveness  of NLRP and traditional teleoperation systems. We also plan to expand the size of the NLRP dictionary and incorporate more primitive skills such as opening/closing cabinet doors and pressing buttons. Additionally, we aim to integrate higher-level instructions, where the robot can convert high-level orders into actions using ChatGPT to find the necessary steps and NLRP to execute them. 

\section{Conclusion}
In conclusion, this paper presented a novel grammar-based natural language framework for converting high-level human text or verbal commands into robot motion for pick-and-place tasks. An initial user study validated the framework's effectiveness and usability, with users reporting nominal cognitive load and a high system usability score. The advantage of this approach is that the framework's dictionary can be easily extended without the need for transfer learning-based training or large datasets. Overall, the presented framework has promising potential for enabling more intuitive and efficient human-robot interactions in a variety of real-world applications.
% In this article, we presented a grammer-based natural language framework that is used to convert the high-level human text commands or verbal commands to respective robot motion. We validated the presented framework with an initial user study and found that the use of the framework for pick and place task imparts nominal cognitive load on users with a high system usability score. The advantage of this grammer-based approach is that the framework's dictionary can easily be extended without relying any transfer learning-based training or large data sets. In the future, we plan to compare the presented framework with different approaches of human-assisted pick and place tasks such as teleoperation-based pick and place with haptic feedback in challenging scenarios.   

\bibliographystyle{ieeeconf}
\bibliography{references}

\end{document}